\documentclass[letterpaper]{article} 
\usepackage{aaai24}  
\usepackage{times}  
\usepackage{helvet}  
\usepackage{courier}  
\usepackage[hyphens]{url}  
\usepackage{graphicx} 
\usepackage{natbib}  
\usepackage{caption} 
\usepackage{amsthm}
\usepackage{amsmath}
\newtheorem{theorem}{Theorem}
\newtheorem{definition}{Definition}
\usepackage[inkscapelatex=false]{svg}
\usepackage[ruled,boxed]{algorithm2e}
\makeatletter 
\g@addto@macro{\@algocf@init}{\SetKwInOut{Parameter}{Parameter}} 
\makeatother
\usepackage{amsfonts}
\usepackage{bbding}
\usepackage{booktabs}

\usepackage{newfloat}
\usepackage{listings}
\DeclareCaptionStyle{ruled}{labelfont=normalfont,labelsep=colon,strut=off} 
\title{H-ensemble: an Information Theoretic Approach to \\Reliable Few-Shot Multi-Source-Free Transfer}
\author {
    Yanru Wu \textsuperscript{\rm 1}, 
    Jianning Wang \textsuperscript{\rm 2},
    Weida Wang \textsuperscript{\rm 1},
    Yang Li \textsuperscript{\rm 1}\footnote{Corresponding author.} 
}
\affiliations {
    \textsuperscript{\rm 1}Tsinghua Shenzhen International Graduation School, Tsinghua University\footnote{Yanru Wu, Weida Wang and Yang Li are from the Shenzhen Key Laboratory of Ubiquitous Data Enabling, SIGS, Tsinghua.}\\
    \textsuperscript{\rm 2}School of Computer Science and Engineering, Harbin Institute of Technology, Shenzhen\\
    \{wu-yr21, wangwd19\}@mails.tsinghua.edu.cn, wangjianning@stu.hit.edu.cn, yangli@sz.tsinghua.edu.cn
}

\begin{document}

\maketitle

\begin{abstract}
Multi-source transfer learning is an effective solution to data scarcity by utilizing multiple source tasks for the learning of the target task. However, access to source data and model details is limited in the era of commercial models, giving rise to the setting of \textit{multi-source-free} (MSF) transfer learning that aims to leverage source domain knowledge without such access. As a newly defined problem paradigm, MSF transfer learning remains largely underexplored and not clearly formulated. In this work, we adopt an information theoretic perspective on it and propose a framework named \textit{H-ensemble}, which dynamically learns the optimal linear combination, or ensemble, of source models for the target task, using a generalization of maximal correlation regression. The ensemble weights are optimized by maximizing an information theoretic metric for transferability. Compared to previous works, H-ensemble is characterized by: 1) its adaptability to a novel and realistic MSF setting for few-shot target tasks, 2) theoretical reliability, 3) a lightweight structure easy to interpret and adapt. Our method is empirically validated by ablation studies, along with extensive comparative analysis with other task ensemble and transfer learning methods. We show that the H-ensemble can successfully learn the optimal task ensemble, as well as outperform prior arts. 

\end{abstract}

\section{Introduction}

The scarcity of annotated data is a principal challenge to machine learning algorithms in a variety of real-world scenarios, where transfer learning has emerged as an effective remedy, harnessing information and insights from well-established source tasks to augment data-deficient domains \citep[see][]{perkins1992transfer, weiss2016survey, zhuang2020comprehensive}. Historically, transfer learning has centered on a singular source-task and target-task relationship. However, the spotlight is gradually shifting towards multi-source transfer learning, capitalizing on multiple source tasks to facilitate the training of the target task \citep[see][]{sun2015survey}. A variety of works have contributed to both the theoretical foundation and practical algorithms on this topic. 
Yet, the trend of decentralization and growing privacy concerns have impeded access to source data, particularly when compared to the models they've nurtured \citep{feng2021kd3a}. This evolving landscape underscores the need for the novel paradigm of \textit{multi-source-free} (MSF) transfer learning \citep{fang2022source}.



\begin{figure}[ht]
    \centering
    \centerline{\includesvg[width=0.8\columnwidth]{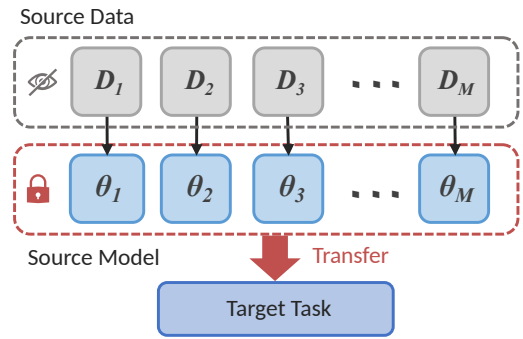}}
    \caption{\textbf{Problem Setting of Multi-Source-Free Transfer Learning.} The transfer to the target task is based on the source models (whose access in our redefined MSF setting is also restricted) trained on the inaccessible source data. }
    \label{fig:setup}
\end{figure}

MSF transfer learning, sometimes referred to as MSF domain adaptation \citep{han2023discriminability}, entails transferring knowledge from multiple sources to a target task without accessing source data, as depicted in Fig. \ref{fig:setup}.
Despite not being clearly defined in most previous papers and still largely underexplored, MSF transfer has been explored by several works these years. Preliminary efforts centered around simple strategies like averaging source models or reducing the problem to single source-free transfers, by source selection using empirical validations or transferability metrics \citep[e.g.][]{yang2023pick}. Contemporary research has pivoted towards sophisticated model ensembles, assigning weights to sources based on perceived importance. For instance, approaches range from correlating source relevance to the target \citep{lee2019learning}, to conditional entropy minimization \citep{ahmed2021unsupervised}, and even leveraging source-similarity and class-relationship perception modules for weight determination \citep{dong2021confident, han2023discriminability}. 
Beyond these linear ensemble techniques, some researchers also introduced auxiliary models to endow cross-domain capabilities \citep{li2022source}.

\subsection{Motivation}

While these methods have their merits, 
they still present three distinct limitations.
Firstly,
they mainly evaluate the source importance as an independent characteristic of each source instead of viewing them as a
whole,
neglecting the likely interaction effects when combining.
Secondly, these works
focus on the unsupervised MSF transfer,
while a more common setting of data scarcity in practice is \textit{few-shot} learning, where we can only access a few of the target data but with labels \citep{wang2020generalizing}. Thirdly, most of them hinge on accessing intermediate layer outputs or necessitate tuning of source models, which tend to become inaccessible in the era of proprietary large models. Also, despite their effectiveness, these transfer algorithms have quite complicated methodologies, making them hard to interpret or adapt. 


As a solution to these issues, we focus on the few-shot MSF transfer challenge, proposing a novel transfer framework tailored to this setting. 
Given the scarcity of target samples and the inaccessibility of source details, we base our methodology on the optimization of features instead of the analysis of input data as other works did.
Specifically, following the idea of linear source ensemble, we define the target feature as a weighted combination of transferred features and introduce an information theoretic metric, \textit{H-score}, to gauge the quality of this synthesis, which can be alternatively interpreted as the transferability of the whole source model ensemble. The optimal source weights will then be simply determined by maximizing this metric, leveraging the robustness of H-score under few-shot settings as well as its computational efficiency and theoretical reliability \citep{huang2019information}.
Our derived target generalizes the maximal correlation regression (MCR) by replacing its feature by an ensemble of source feature extractors and modifying the model accordingly.
A detailed explanation of the underlying theoretic basis and intuition of H-ensemble framework will be presented in the Methodology section, showing that: 1) under the maximal correlation framework, the weights maximizing our adapted H-score are theoretically optimal for the target task; 2) the resulting optimal question is of a benign form and can be reliably solved by gradient descend based algorithms.

\begin{table*}[htb]
\setlength{\tabcolsep}{4pt}
    \renewcommand{\arraystretch}{1.2}
    \centering
    \begin{tabular}{c c c c c}
        \toprule
         \textbf{Setting} &  \textbf{Multi-Source} & \textbf{No Source Data} & \textbf{No Source Model Details} & \textbf{Target} \\
         \midrule
         Few-Shot transfer \citep[]{wang2020generalizing} & \XSolidBrush & \XSolidBrush & \XSolidBrush & Few-shot \\
         Unsupervised DA \citep[]{hoffman2018cycada} &  \XSolidBrush & \XSolidBrush & \XSolidBrush & Unsupervised \\
         Multi-Source DA \citep[]{peng2019moment} &  \Checkmark & \XSolidBrush & \XSolidBrush & Unsupervised \\
         Multi-Source-Free DA \citep[]{ahmed2021unsupervised}  &  \Checkmark & \Checkmark & \XSolidBrush & Unsupervised \\
         Few-Shot Multi-Source-Free Transfer (Ours)  &  \Checkmark &  \Checkmark &  \Checkmark & Few-shot \\
         \bottomrule
    \end{tabular}
    \caption{\textbf{Comparison across Relative Problem Settings,} involving the aspects of whether there are multi-sources, no source data, no source model details, and different target learning approaches. The abbreviation `DA' represents domain adaptation, `\Checkmark' represents obtaining the corresponding aspects, while `\XSolidBrush' the opposite. Among all settings, ours is relatively restrictive.}
    \label{tab:setting}
\end{table*}


\subsection{Contributions}
In this paper, we propose an information theoretic MSF transfer framework named \textit{H-ensemble}, where the source models are assembled using a generalization of MCR, and the source weights are obtained by maximizing an H-score based metric. 
Our contributions mainly lie in three aspects:

\begin{itemize}
    \item[a)] We formulate the few-shot MSF transfer problem mathematically using a source-free setting that is more common in practical scenarios.
    
        
    \item[b)] An efficient and theoretically reliable framework derived from maximal correlation regression is introduced for MSF transfer. The framework is designed to be lightweight and easy to interpret and adapt, possible to serve as a \textit{plug-in} module. 
    
    \item[c)] We present both theoretical guarantee and experimental support to the effectiveness of our methodology.

    
\end{itemize}

\section{Related Works}
\subsection{Maximal Correlation Regression and H-Score}\label{sec:H-score}

 Maximal correlation regression (MCR) network \citep{xu2020maximal} refers to a specially designed network trained based on the optimization of feature quality by maximizing H-score. Here, H-score is a metric determining whether certain features are informative for the task  \citep[more formally introduced by][]{bao2019information}\footnote{The H-score introduced by Bao et al. is actually the one-sided form of the score in MCR (both firstly defined by Huang et al. in 2019). They are also differed in their definition of variables and optimal objects.}. Its derivation is rooted in the maximal correlation interpretation of deep neural networks \citep{huang2019universal}, where given the extracted features, the formula of H-score is actually the optimal network performance under an information theoretic measurement. Previous work also extended H-score to a transferability metric and verified its effectiveness with extensive experiments \citep{bao2019information, ibrahim2022newer}, suggesting the potential of H-score in transfer learning.

Notably, as the basis of both MCR network and H-score, maximal correlation analysis originates from the works of Hirschfeld, Gebelein and Renyi \citep{hirschfeld1935connection, gebelein1941statistische, renyi1959measures} and has been followed and further explored by a broad spectrum of successive work. A detailed overview of maximal correlation analysis can be found given by \citeauthor{huang2019universal} (\citeyear{huang2019universal}).




\subsection{Multi-Source Transfer Learning}\label{sec:multisource}



Multi-source transfer learning, a well-established concept, encompasses approaches like naive multi-source transfer by feature extractor concatenation \citep{christodoulidis2016multisource} and theoretical explorations \citep{crammer2008learning, mansour2008domain, ben2010theory, tong2021mathematical, chen2023algorithm}. Contemporary algorithms can be classified into three main categories: re-weighting approaches that estimate and assign source importance \citep{sun2011two, lee2019learning, shui2021aggregating}; source selection methods using transferability metrics to identify optimal combinations \citep{agostinelli2022transferability}; and boosting-based solutions that apply ensemble learning techniques to transfer learning \citep{huang2012boosting, xu2012multi, fang2019adapted}. Further insights on this topic are available in related surveys, such as \citep{sun2015survey}.

\section{Problem Definition}

\subsection{Setting Comparison to Previous Works}

Based on the change of settings mentioned in the Introduction section, there are mainly three terms to emphasize in our setting: multi-source, source-free and few-shot. Specially, to distinguish from previous notions, source-free is categorized into source-data-free and source-model-free. We compare few-shot MSF transfer with previous settings in these four aspects and summarize the main divergence in Table~\ref{tab:setting}. 


 \begin{table}[!ht]
     \renewcommand{\arraystretch}{1.2}
    \centering
    \begin{tabular}{c |c }
        \toprule
        \textbf{Notation} & \textbf{Meaning}\\
        \midrule
        $S_j$; $T$ & Source/Target Task\\
        $d_{S_j}$; $d_T$& Source/Target Class Number\\
     $N_{S_j}$; $N_T$ & Source/Target Data Size \\
         $D_{S_j} = \{(x_{S_j}^i, y_{S_j}^i)\}_{i = 1}^{N_{S_j}}$ & Source Data \\
         $D_T = \{(x_T^i, y_T^i)\}_{i = 1}^{N_T}$ & Target Data \\
          \bottomrule
    \end{tabular}
    \caption{\textbf{Notation.} We omit $\{\cdot\}_{j=1}^M$ for simplification.}
    \label{tab:notation}
\end{table}

\subsection{Mathematical Formulation}

Consider the transfer learning setting with $M$ source tasks $\{S_j\}_{j=1}^M$ and a target task $T$, we list the main notations in Table~\ref{tab:notation}.  For simplification, we assume that the sources and target tasks share the same input space $X$ and all tasks are classification problems.\footnote{This assumption is without loss of generality, for in most cases,
model inputs can be mapped to any size by preprocessing (e.g. image resizing, pooling, certain embedding layers, etc.), 
and classification is one of the most representative tasks in machine learning.}

Notably, in our setting we have no access to source data $D_{S_j}$ and only partial access to target data $D_T$, so here $N_{S_j} = 0$, $N_T = k \times d_T$ for a $k$-shot MSF transfer.\footnote{$k$ samples for each of the $d_T$ classes} 
For the source model consisting of a backbone feature extractor $\boldsymbol{f}_{S_j}: X \to \mathbb{R}^{d_f}$ and a classification head $\boldsymbol{g}_{S_j}: \mathbb{R}^{d_f} \to \mathbb{R}^{d_{S_j}}$ ($d_f$ denotes the feature dimension)\footnote{$d_f$ is set to the same across all tasks considering that we could add a fully connected layer to derive any output dimension.}, where the source-free restriction leaves the detailed model structure unknown.


In computation, we estimate the distribution $P^T_{X}$, $P^T_{Y}$ of target data $x_T$, $y_T$ with empirical distribution $\hat P^T_{X}$, $\hat P^T_{Y}$:

\begin{equation}
\hat P_{X}^T (x)= \frac{1}{N_T}\sum_{i=1}^{N_T}\mathbb{I}\{x_T^i=x\}; \hat P_{Y}^T (y)= \frac{1}{N_T}\sum_{i=1}^{N_T}\mathbb{I}\{y_T^i=y\}, 
\label{py}
\end{equation}

\noindent where $\mathbb{I}\{\cdot\}$ is the indicator function \citep{feller1991introduction}. Similarly, the conditional distribution of $x_T$ given $y_T$ $P^T_{X|Y}$ is estimated by its empirical distribution $\hat P^T_{X|Y}$:

\begin{equation}
\hat P^T_{X|Y} (x|y)= \frac{\sum_{i=1}^{N_T}\mathbb{I}\{x_T^i=x,y_T^i=y\}}{\sum_{i=1}^{N_T}\mathbb{I}\{y_T^i=y\}}.   
\label{pxy}
\end{equation}

When estimating the underlying distributions with empirical distributions of samples, Eqns. (\ref{py}) and (\ref{pxy}) may be poorly defined and intractable for most target task types due to the high dimension of input $x$. However, this problem can be avoided, as in our framework we only need to compute the expectation of features over these distributions.

The goal of few-shot MSF problem is that, by utilizing the partially accessible data and black-box source models only, find the optimal parameters $\boldsymbol{\theta}_T$ that minimizes certain divergence $\mathcal L$ between the real conditional distribution $P^T_{Y|X}$ and the approximated $P^{(\boldsymbol{\theta}_T)}_{Y|X}$, \textit{i.e.},

\begin{equation}
    \boldsymbol{\theta}^*_T = arg\min_{\boldsymbol{\theta}_T} \mathcal L(P^{(\boldsymbol{\theta}_T)}_{Y|X}, P^T_{Y|X}).
    \label{definition}
\end{equation}

\begin{figure*}
    \centering
    \centerline{\includesvg[width=2\columnwidth]{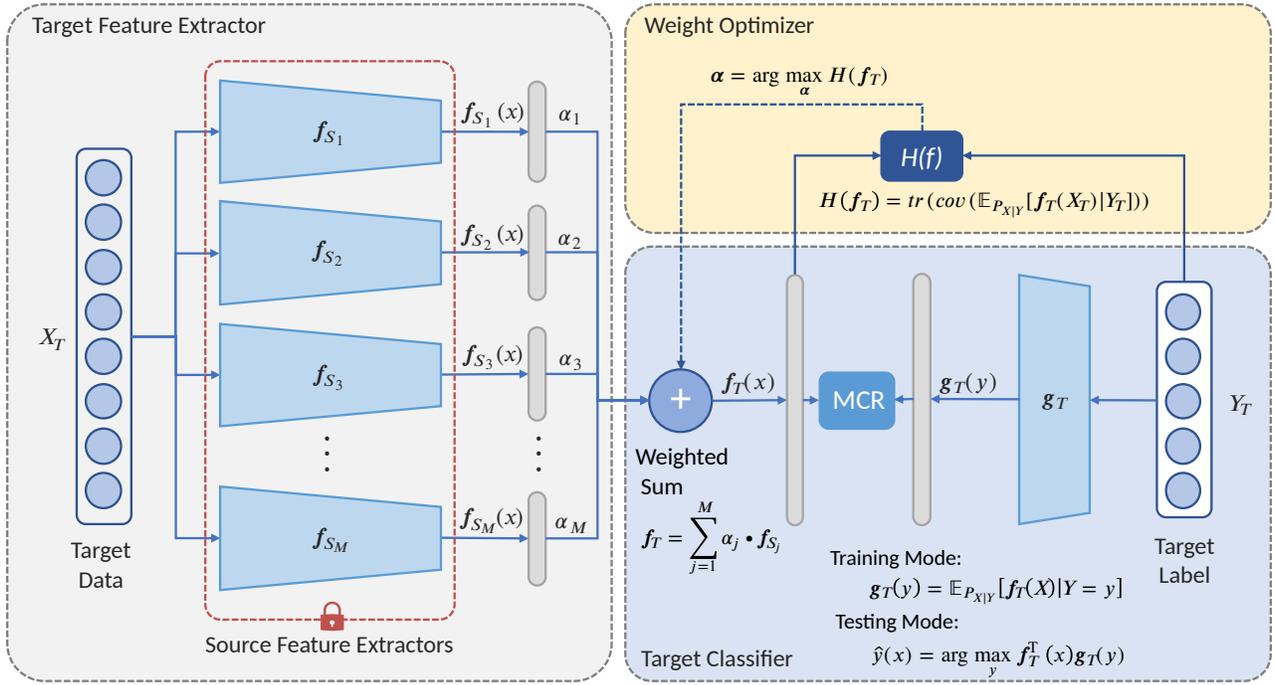}}
    \caption{\textbf{Framework of H-Ensemble.} The framework consists of three modules. The target data firstly flow into Target Feature Extractor. Then the Weight Optimizer will utilize the outputs of source feature extractors and target label to derive the optimal source weight $\boldsymbol{\alpha}$, which makes the parameter in deriving target feature. Finally the Target Classifier will be trained and used together with the extractor for test according to a generalization of maximal correlation regression (MCR).}
    \label{fig:framework}
\end{figure*}

\section{Methodology}

Following the analysis in the Motivation section, we model the target feature extractor $\boldsymbol{f}_T$  by a weighted combination of source feature extractors $\boldsymbol{f}_{S_j}$: 

\begin{equation}
    \begin{aligned}
        \boldsymbol{f}_T = \sum_{j=1}^M \alpha_j \cdot \boldsymbol{f}_{s_j}, \ \sum_{j=1}^M \alpha_j = 1 .
    \end{aligned}
    \label{targetfeature}
\end{equation}

We then need to efficiently evaluate target features to guide the optimization of weights $\boldsymbol{\alpha}$, where the maximal correlation analysis (MCA) is introduced. In MCA, a neural network is viewed as a feature extractor $\boldsymbol{f}$ and a classifier $\boldsymbol{g}$, and the informativeness of features extracted by $\boldsymbol{f}$ can be estimated by H-score. 

Leveraging the robustness of H-score under few-shot settings as well as its computational efficiency and theoretical reliability, we use H-score as the feature quality measurement and set up the transfer framework under the same theory scheme as a generalization of MCR. Following previous works \citep{xu2020maximal}, the divergence $\mathcal L$ in Eqn. (\ref{definition}) is defined as variational chi-squared divergence: 
\begin{equation}
    \mathcal L(\boldsymbol{\theta}) = \sum_x P_X(x) \sum_y \frac{[P^{(\boldsymbol{\theta})}_{Y|X} (y|x) - P_{Y|X} (y|x)]^2}{P_Y(y)}.  
\end{equation}

We illustrate our MSF transfer framework H-ensemble in Fig. \ref{fig:framework} and introduce each module in the sections below, showing how the divergence $\mathcal L$ is minimized.

\subsection{Multi-Source-Free Transfer}


Overall, the MSF transfer framework follows the generalized MCR\citep{xu2020maximal} network and has two modules, feature extractor $\boldsymbol{f}_T = \sum_{j=1}^M \alpha_j \cdot \boldsymbol{f}_{s_j}$ and classifier $\boldsymbol{g}_T$ derived using MCR training mode as below. The approximated conditional distribution is defined as 


\begin{equation}
    P^{(\boldsymbol{\theta_T})}_{Y|X} (y|x) = P^T_Y(y) (1+\boldsymbol{f}_T^{\rm T}(x)\boldsymbol{g}_T(y)).
\end{equation}


Specifically, given input data $x \sim P^T_X$, the features extracted by source models are $\boldsymbol{f}_{s_j}(x)$. To simplify the deduction, we assume that $\mathbb{E}_{P^T_X}[\boldsymbol{f}_{s_j}(x)] = 0$ and $\mathbb{E}_{P^T_X}[\boldsymbol{f}_{s_j}(x)\cdot \boldsymbol{f}_{s_j}(x)^{\rm T}] = \mathbf{I}$ for all $j$, as a normalization layer could be efficiently added to extractors. Subsequently, $\boldsymbol{f}_T$ is also a normalized feature with $\mathbb{E}_{P^T_X}[\boldsymbol{f}_{T}(x)] = 0$ and $\mathbb{E}_{P^T_X}[\boldsymbol{f}_{T}(x)\cdot \boldsymbol{f}_{T}(x)^{\rm T}] = \mathbf{I}$. After deriving the target feature extractor $\boldsymbol{f}_T$, the optimal target classifier $\boldsymbol{g}_T$ minimizing $\mathcal L(\boldsymbol{\theta}_T)$ can be directly calculated by generalizing MCR to a multiple extractor combination scenario. 

\begin{theorem}
    Given feature extractors of function form $\boldsymbol{f}_{i}(x)$, $i \in \{1,\dots, n\} $ and the linearly combined feature extractor $\boldsymbol{f}(x) = \sum_{i=1}^n \alpha_i \cdot \boldsymbol{f}_i(x)$, where $\mathbb{E}_{P_X}[\boldsymbol{f}_i(x)] = 0$, $\mathbb{E}_{P_X}[\boldsymbol{f}_i(x)\cdot \boldsymbol{f}_i(x)^{\rm T}] = \mathbf{I}, \forall i$ and $\sum_{i=1}^n \alpha_i = 1$. The optimal classifier $\boldsymbol{g}^*(y)$ will be of the form:
    \begin{equation}
    \begin{aligned}
        \boldsymbol{g}^*(y) =   \sum_{i=1}^n\ \  & \alpha_i \cdot \mathbb{E}_{P_{X|Y}}[\boldsymbol{f}_i(X)|Y=y],  
    \end{aligned}
    \label{gnet}
    \end{equation}
    \label{getg}
    where the proof is provided in the Appendix.I. 
\end{theorem}

By using Eqn. (\ref{gnet}), we can 
skip training and directly set the target classifier $\boldsymbol{g}_T(y)$ to the theoretical optimal $\boldsymbol{g}_T^*(y) = \sum_{i=1}^n\alpha_i \cdot \mathbb{E}_{P_{X|Y}}[\boldsymbol{f}_i(X)|Y=y]$, thereby achieving high efficiency in transfer. 

Notably, the form of $\boldsymbol{g}_T(y)$ does not conform to the usual classifiers that take in extracted features and output the class probabilities (as in the source classifiers $\boldsymbol{g}_{S_j}: \mathbb{R}^{d_f} \to \mathbb{R}^{d_{S_j}}$). On the contrary, the target classifier $\boldsymbol{g}_T(y)$ 
acts as a label encoder, 
which maps labels $y$ to a newly defined y-feature, \textit{i.e.} $\boldsymbol{g}_T: \mathbb{R}^{d_T} \to \mathbb{R}^{d_{f}}$. In testing mode, the classification result is given by maximizing the correlation between the features extracted, \textit{i.e.} $\hat{y}(x) = arg\max_y \boldsymbol{f}^{\rm T}_T(x) \boldsymbol{g}_T(y)$. More details of this network architecture are discussed in previous works \citep{huang2019universal}.

\begin{algorithm*}[!h]
\KwIn{Target data $D_T = \{(x_T^i, y_T^i)\}_{i = 1}^{N_T}$, source feature extractors $\{\boldsymbol{f}_{S_j}\}_{j = 1}^{M}$}
\Parameter{Learning rate $\lambda$}
\KwOut{Target classifier $\boldsymbol{g}_T$, source weight $\boldsymbol{\alpha}$}
    Randomly initialize $\boldsymbol{\alpha}=\{\alpha_1,\alpha_2,\dots,\alpha_M\}$, $\ \ \sum_{j=1}^M \alpha_j = 1$ \;
    \For(\tcp*[f]{Compute empirical conditional expectation}){$y \gets 1$ \KwTo $d_T$}{
          $n_y \gets \sum_{i = 1}^{N_T} \mathbb{I}\{y = y^i_T\}$ \;
          $\mathbb{E}_{P^T_{X|Y}} [\boldsymbol{f}_{S_j}(X_T)|Y_T = y] \gets \frac{1}{n_y} \sum_{i = 1}^{N_T} \boldsymbol{f}_{S_j}(x_T^i) \cdot \mathbb{I}\{y = y^i_T\}$, $\ \ j = 1,\dots,M$
          }
    \Repeat(\tcp*[f]{Optimize weight}){$\alpha$ converges}{
          
          $H(\boldsymbol{f}_T;\boldsymbol{\alpha}) \gets tr(cov(\sum_{j=1}^M \alpha_j \mathbb{E}_{P^T_{X|Y}}[\boldsymbol{f}_{S_j}(X_T)|Y_T]))$\;
          $\boldsymbol{\alpha} \gets \boldsymbol{\alpha} + \lambda \nabla_{\boldsymbol{\alpha}} H(\boldsymbol{f}_T; \boldsymbol{\alpha})$\;
          \For(\tcp*[f]{Project weight to hyperplane $\sum_{j=1}^M \alpha_j = 1$}){$j \gets 1$ \KwTo $M$}{
          $\alpha_j \gets \alpha_j - \frac{1}{M} \sum_{j=1}^M \alpha_j + \frac{1}{M}$
          }
        }
    \For(\tcp*[f]{Compute classifier function}){$y \gets 1$ \KwTo $d_T$}{
          $\boldsymbol{g}_T(y) =   \sum_{j=1}^M\ \alpha_j \cdot \mathbb{E}_{P_{X|Y}}[\boldsymbol{f}_{S_j}(X_T)|Y_T=y] $
          }
\caption{H-ensemble: Training}
\label{htrain}
\end{algorithm*}


\subsection{H-Score for Estimating Feature Transferability}

 Defining the target feature as $\boldsymbol{f}_T = \sum_{j=1}^M \alpha_j \cdot \boldsymbol{f}_{s_j}$, we introduce the information theoretic metric H-score \citep{huang2019information} to measure the feature transferability for weight optimization. In this section, we give a detailed formulation of H-score as well as its intuitive interpretation. From MCA, the mathematical form of H-score is defined as follows.

\begin{definition}
    With input data $x$, label $y$, feature extractor $\boldsymbol{f}(x)$ and classifier $\boldsymbol{g}(y)$ (both zero-mean functions). The H-score of $\boldsymbol{f}$ with regard to the task casting $x$ to $y$ is:
    \begin{align}
        &H(\boldsymbol{f},\boldsymbol{g}) = \nonumber\\
        &\mathbb{E}_{P_{XY}}\left[\boldsymbol{f}^{\rm T}(X)\boldsymbol{g}(Y)\right] - \frac{1}{2}tr(cov(\boldsymbol{f}(X)) cov(\boldsymbol{g}(Y))).
    \label{hscore}
    \end{align}
\label{def:hscore}    
\end{definition}

Since we have defined our classifier $\boldsymbol{g}_T(y)$ as the optimal classifier $\boldsymbol{g}_T^*(y)$ in the previous section, the H-score can be simplified to the one-sided H-score.

\begin{definition}
    With input data $x$, label $y$ and feature extractor $\boldsymbol{f}(x)$ (a zero-mean feature function). The one-sided H-score of $\boldsymbol{f}$ with regard to the task casting $x$ to $y$ is:
    \begin{equation}
        H(\boldsymbol{f}) = tr(cov(\boldsymbol{f}(X))^{-1}cov(\mathbb{E}_{P_{X|Y}}[\boldsymbol{f}(X)|Y])).
    \label{singlehscore}
    \end{equation}
 \label{def:singlehscore}
\end{definition}

\begin{theorem}
    For a task T with feature extractor $\boldsymbol{f}$ and classifier $\boldsymbol{g}$, the minimization of $\mathcal L$ in maximal correlation regression is equivalent to the maximization of H-score, \textit{i.e.}
    \begin{equation}
         \boldsymbol{f}^*, \boldsymbol{g}^* = arg \min_{\boldsymbol{f},\boldsymbol{g}}\mathcal L = arg \max_{\boldsymbol{f},\boldsymbol{g}} H(\boldsymbol{f}, \boldsymbol{g}).
    \end{equation}
     \noindent Moreover, if $\boldsymbol{g}$ is defined as the optimal classifier $\boldsymbol{g}^*$, $H(\boldsymbol{f},\boldsymbol{g})$ is then equal to the one-sided H-score $H(\boldsymbol{f})$,
    \textit{i.e.}
    \begin{equation}
    H(\boldsymbol{f}, \boldsymbol{g}^*) = H(\boldsymbol{f}) .
    \label{hscoremeaning}
    \end{equation}
    \label{theo:hscore}
\end{theorem}

Theorem~\ref{theo:hscore} indicates that the minimization of $\mathcal L(\boldsymbol{\theta})$ is equivalent to the maximization of the one-sided H-score $H(\boldsymbol{f})$, and therefore we can determine the optimal target feature in our method by simply maximizing $H(\boldsymbol{f}_T)$. 

The formula of one-sided H-score in Eqn. (\ref{singlehscore}) can be intuitively interpreted as normalizing the inter-class feature variance $cov(\mathbb{E}_{P_{X|Y}}[\boldsymbol{f}(X)|Y])$ with feature redundancy $tr(cov(f(X)))$, reflecting the efficiency of extracted feature in distinguishing different classes. The proof of Eqn. (\ref{hscore}), Eqn. (\ref{singlehscore}) and Theorem \ref{theo:hscore} can be found in \citep{huang2019information, bao2019information,xu2020maximal}. Considering that our target feature $\boldsymbol{f}_T$ is normalized, we can further simplify Eqn. (\ref{singlehscore}) to:

\begin{equation}
    H(\boldsymbol{f}_T) = tr(cov(\mathbb{E}_{P_{X|Y}^T}[\boldsymbol{f}(X_T)|Y_T])).
    \label{nhscore}
\end{equation}

 Eqn. (\ref{nhscore}) can be explicitly computed from extracted features and labels. Hence, using target data $D_T$, we could efficiently estimate in advance the transferred performance, or transferability, of target feature $\boldsymbol{f}_T(x)$, and optimize the source weights accordingly. 
For clarity, in the following sections we refer to the score defined in Eqn. (\ref{nhscore}) as `H-score'.



\subsection{Source Feature Re-weighting}

With the source feature extractors and target samples fixed, the target feature $\boldsymbol{f}_T$ can be viewed as a function of the weight $\boldsymbol{\alpha}$ defined in Eqn. (\ref{targetfeature}) and therefore the same for its H-score $H(\boldsymbol{f}_T)$. Hence, by maximizing $H(\boldsymbol{f}_T; \boldsymbol{\alpha})$ with respect to $\boldsymbol{\alpha}$, we will be able to determine the optimal weight $\boldsymbol{\alpha}$ for minimizing the loss $\mathcal L(\boldsymbol{\theta})$ in multi-source transfer, alternately interpreted as deriving the target feature of the highest transferability. We formulate our optimization problem as follows.

\begin{definition}
    With input data $x_T$, label $y_T$ and feature extractors $\boldsymbol{f}_{S_j}(x)$ for $j \in \{1, \dots, M\}$ (all zero-mean, unit-variance feature functions), the optimal feature weight $\boldsymbol{\alpha} = (\alpha_1, \alpha_2, \dots, \alpha_M)^T \in \mathbb{R}^M$
    will be given by:
    \begin{equation}
        \begin{aligned}
        \boldsymbol{\alpha}^* = arg\max_{\boldsymbol{\alpha}}  H\left(\sum_{j=1}^M \alpha_{j} \cdot \boldsymbol{f}_{S_j}\right)  \ 
        s.t.\ \sum_{j=1}^M \alpha_j = 1 .
        \end{aligned}
    \label{tscore}    
    \end{equation}
 \label{def:tscore}
\end{definition}

The linear constraint of $\boldsymbol{\alpha}$ is obviously convex. We then verify the benign property of the proposed optimization problem by proving that the optimal object H-score is also a convex function of $\boldsymbol{\alpha}$.

\begin{theorem}
      With input data $x$ and label $y$, when weighted $\alpha_i$ summing up to 1 and fixed features $\boldsymbol{f}_i$ being zero-mean, unit-variance ($i \in \{1, \dots, n\}$), the H-score of the weighted feature sum will be a convex quadratic form of $\boldsymbol{\alpha}$ as below:
    \begin{equation}
    \begin{aligned}
     &H(\boldsymbol{f}) = H\left(\sum_{i=1}^n \alpha_i \cdot \boldsymbol{f}_i\right) = \sum_{i=1, j=1}^{n,n}\alpha_i \alpha_j \\
     &\cdot tr(\mathbb{E}_{P_{Y}}[\mathbb{E}_{P_{X|Y}}[\boldsymbol{f}_i(X)|Y]\cdot \mathbb{E}_{P_{X|Y}}[\boldsymbol{f}_j(X)|Y]^T])   .
    \end{aligned}
    \end{equation}
    \label{theo:convex}
\end{theorem}

 \begin{table*}[!h]
    \renewcommand{\arraystretch}{1.2}
    \centering
    \begin{tabular}{c |c c c c| c c c c| c }
          \toprule
          Method & R $\to$ t0 & R $\to$ t1 & R $\to$ t2 & R $\to$ t3 & R $\to$ v0 & R $\to$ v1 & R $\to$ v2 & R $\to$ v3 & Avg. \\
          \midrule

TargetOnly & 0.9730 & 0.9675 & 0.9165 & 0.7260 & 0.9180 & 0.8285 & 0.8750 & 0.7645 & 0.8711 \\
          Single-Average & 0.9697 & 0.9659 & 0.9266 & 0.7503 & 0.8416 & 0.7640 & 0.7753 & 0.7013 & 0.8368 \\
          Single-Best & 0.9830 & 0.9725 & \underline{0.9405} &  \textbf{0.7760} & 0.9435 &  \textbf{0.8470} & \underline{0.8915} &  \underline{0.8110} & \underline{0.8956}\\
          \midrule
	Average W. & 0.9840 & \textbf{0.9770} & 0.9340 & 0.7510 & 0.9240 & 0.8370 & 0.8355 & 0.7910 & 0.8792 \\

    MultiFinetune  & 0.9695 & 0.9715 & 0.9400 &  0.7570 & 0.9105 &  0.8150 & 0.8240 &  0.7760 & 0.8704
 \\
           \midrule
	MCW & \textbf{0.9890} & 0.9585 & \textbf{0.9415} & 0.7675 & \underline{0.9460} & 0.8160 & 0.8655 & 0.7600 & 0.8805 \\
	DECISION & 0.5590 & 0.4635 & 0.7700 & 0.3125 & 0.3055 & 0.3245 & 0.5435 & 0.3040 & 0.4478 \\
	DATE & 0.2490 & 0.3655 & 0.7740 & 0.2635 & 0.2500 & 0.4545 & 0.5155 & 0.1260 & 0.3748 \\
          \midrule
H-ensemble* & \underline{0.9855} & \underline{0.9745} & \textbf{0.9415} & \underline{0.7690} & \textbf{0.9470} & \underline{0.8420} & \textbf{0.8990} & \textbf{0.8300} & \textbf{0.8986} \\
          \bottomrule
    \end{tabular}

    \caption{\textbf{Accuracy Comparison on VisDA-2017 (10-shot).} `R' stands for the rest tasks. The highest/second-highest accuracy is marked in \textbf{Bold}/\underline{Underscore} form respectively. Our method achieves the overall best performance.
    }
    \label{tab:comparison}
\end{table*}

 A detailed proof of Theorem \ref{theo:convex} is presented in the Appendix.I. The optimization problem of $\boldsymbol{\alpha}$ can therefore be reliably solved by gradient descent (GD) based algorithms with theoretical guarantee. We then use the optimal $\boldsymbol{\alpha}$ and derive the optimal target model as defined in Eqn. (\ref{definition}) with Eqn. (\ref{targetfeature}) and (\ref{gnet}).
 

As a possible solution, we use projected gradient descent (PGD) for optimization and present the resulting training process of H-ensemble in Algorithm \ref{htrain}. The derivation of the projection formula in PGD is included in Appendix.I. The predicted result in testing will be efficiently computed by:
\begin{equation}
    \begin{aligned}
            \hat{y}(x) &= arg\max_y \left(\sum_{j = 1}^{M} \alpha_j \boldsymbol{f}_{S_j}^{\rm T}(x)\right)\cdot\boldsymbol{g}_T(y).
    \end{aligned}
\end{equation}


\section{Experimental Results}\label{exp}

\subsection{Experiment Setup}

\subsubsection{Datasets.}

To verify the effectiveness of H-ensemble in the few-shot MSF setting, we conduct extensive experiments on four benchmark datasets. \textbf{VisDA-2017} \citep{peng2017visda} is a visual domain transfer challenge dataset containing over 280,000 images across 12 categories in training (T, synthetic) and validation (V, real) domains respectively.
\textbf{Office-31} \citep{saenko2010adapting} is a standard transfer learning dataset with 4,652 images and 31 unbalanced object categories in three domains: Amazon, DSLR and Webcam, consisting of objects commonly encountered in office scenarios. Developed from Office-31, \textbf{Office-Caltech} \citep{gong2012geodesic} contains 2,533 images covering four domains: Amazon, DSLR, Webcam and Caltech256, each with 10 categories. \textbf{Office-Home} \citep{venkateswara2017deep} is a more challenging dataset with 15,599 images in 65 unbalanced categories collected from four domains: Artistic, Clip Art, Product, and Real-world. 
\footnote{For an evaluation on large-scale datasets, we also compared H-ensemble against other methods on \textbf{DomainNet} \citep{peng2019moment} during rebuttal. The results are included in the Appendix.III.}

\subsubsection{Task Setting.}


We split each domain into several tasks. For example, in VisDA-2017 we further divide both domain V and T into 4 tasks (\textit{i.e.} v0 - v3 and t0 - t3), each with 3 classes. The resulting eight tasks, encompassing both domain and label differences, constitute a diverse task pool for the transfer experiment. 
We do the same on the Office series datasets and the details can be found in the Appendix.II.





Following the standard protocol of few-shot learning, the training data for $k$-shot is $k$ samples per class randomly selected from the target task. Source models are set to Resnet18 with hidden dim 256 trained on full training sets. More details of extra dataset, experimental setup and model implementation are discussed in the Appendix.II.

\subsection{Comparison with Baselines}

\subsubsection{Baselines.}
For a general performance evaluation, we conduct comparison experiments using the first task synthesis strategy. Considering very few methods are designed for few-shot MSF transfer setting, we take SOTA works under similar settings and variations of our method as baselines. The compared methods include: 1) \textbf{Trivial Solutions}: Target Only, Single-Best, Single-Avg.; 2) \textbf{Relevant Approaches}: Average W., MultiFinetune; 3) \textbf{Prior SOTAs}: MCW \citep{lee2019learning}, DECISION \citep{ahmed2021unsupervised}, DATE \citep{han2023discriminability}.
Here, 
Single-* adapt one source model to target task using MCR and *-Best/Avg. stand for the best/average performance. Target Only finetunes Imagenet-pretrained model on few-shot target data only.
Average W. replace the weighted sum in our framework with
equal weights. MultiFinetune learns a fully connected classifier with the target extractor learned in H-ensemble.
The Prior SOTAs are the SOTAs with the closest problem setting (MSFDA). 




\subsubsection{Results and Analysis.}

We take VisDA-2017 10-shot as an example\footnote{Experiments on other datasets \& shots and statistical results are listed in Appendix.III.} and summarize the results in Table~\ref{tab:comparison}. Overall,  H-ensemble performs the best among all the methods including the best single source transfer, especially in the more challenging real world image (V) domain. Here the significant degradation of DECISION and DATE is due to the change of problem setting from UDA to FS transfer. 


\begin{figure}[ht]
    \centering
    \centerline{\includesvg[width=0.95\columnwidth]{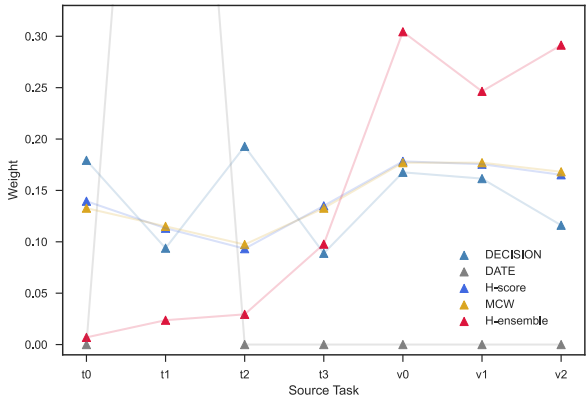}}
    \caption{\textbf{Source Weights derived in Different Methods.} (VisDA-2017, target v3, 10-shot) The source weight in H-ensemble (red) conforms most to intuition, emphasizing on tasks from the same domain (v0, v1, v2).}
    \label{fig:weight}
\end{figure}

For further analysis, we plot the source weights for task v3 derived in weighting-based methods in Fig. \ref{fig:weight}. It is clearly illustrated that given four tasks on domain V and T each, H-ensemble can successfully recognize the importance of integrating tasks from the same domain (v0, v1, v2) and attach them with high weights. It also correctly emphasizes on t3 (the only task sharing the same class labels with v3) the most among tasks on domain T, displaying an outstanding and interpretable ability to measure source importance.


\subsection{Ablation Study}

We further explore the capability of H-ensemble when transferring to harder tasks on Office datasets, as well as verify components in H-ensemble by ablating. The baselines are constructed by substituting each part with naive solutions. Specifically, we have: 
\begin{itemize}
    \item \textbf{Single MCR:} multi-source $\to$ average performance of single-source transfer; 
    \item \textbf{Average W.:} weighting strategy $\to$ average weights; \
    \item \textbf{MultiFinetune:} MCR classifier $\to$ finetune classifier.
\end{itemize}

\begin{figure}[!ht]
    \centering
    \centerline{\includesvg[width=1\columnwidth]{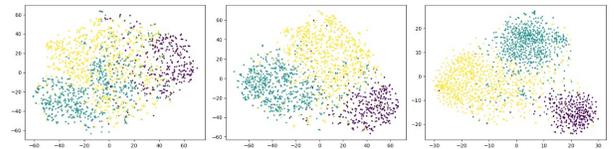}}
    \caption{\textbf{Feature extracted by different models,} visualized by t-SNE. From left to right: random source, Average W. and H-ensemble. It turns out that features generated by our method have the greatest discriminability.
    }
    \label{fig:tsne}
\end{figure}

For each variety, we record the average transfer performance on each dataset, concluding the results of 8-shot\footnote{Compared to VisDA-2017, the shot reduces due to fewer samples per class in Office-* datasets. More results in Appendix.III.} in Table~\ref{tab:ablation}, where both the effectiveness of H-ensemble and the necessity of every module are explicitly supported. It is also notable that our whole framework is established on a united theory ground, and any substitution may break its theoretical reliability and interpretability. We also visualize the target feature extracted by random source, average weight and H-ensemble in Fig. \ref{fig:tsne}, showing that H-ensemble extracts features with the strongest class discriminability for the given task. 


 \begin{table}[!ht]
    \renewcommand{\arraystretch}{1.2}
    \centering
    \begin{tabular}{c |c c c }
          \toprule
          Method & Office-31 & Office-Caltech & Office-Home  \\
          \midrule
          Single MCR & \underline{0.8483}  &  0.9645  &  0.6235 \\
          Average W. &  0.8479  &  \textbf{0.9716}  &  \underline{0.6702} \\
          MultiFinetune  &  0.8414  &  0.9663  &  0.6571 \\
          H-ensemble* & \textbf{0.8644} & \underline{0.9708} & \textbf{0.6940}\\
          \bottomrule
    \end{tabular}
    \caption{\textbf{Ablation Study on Office Datasets (8-shot).} Our weighting strategy and optimal MCR classifier both contribute to the overall effectiveness.
    }
    \label{tab:ablation}
\end{table}

\section{Conclusion}



In this work, we highlight a common yet largely underexplored scenario of multi-source transfer which we refer to as few-shot MSF transfer, and take the lead in giving a mathematical formulation of it. Addressing this problem, we take on an information theoretic perspective and propose our H-ensemble framework. Consisting of three main components, H-ensemble dynamically learns the optimal linear ensemble of source models for the target task, with the ensemble weights optimized by maximizing H-score and classifier determined by a generalization of MCR. We present detailed theoretic deduction and interpretation as well as extensive experimental validation for our method, showing that H-ensemble can effectively boost the learning on the target task in the few-shot MSF transfer scheme.

\section{Acknowledgement}
This study is supported in part by the Tsinghua SIGS Scientific Research Start-up Fund (Grant QD2021012C), Natural Science Foundation of China (Grant 62001266) and Shenzhen Key Laboratory of Ubiquitous Data Enabling (No. ZDSYS20220527171406015).

\bibliography{aaai24}

\end{document}